\documentclass[11pt,a4paper]{article}
\usepackage[hyperref]{naaclhlt2019}
\usepackage{times}
\usepackage{latexsym}
\usepackage{dblfloatfix}

\usepackage{url}
\usepackage{graphicx}
\usepackage{xspace}
\usepackage{pgfplots}
\usepackage{array}
\pgfplotsset{compat=1.14}

\aclfinalcopy % Uncomment this line for the final submission

\newcommand{\obverse}[1]{\href{https://www.openbible.info/labs/cross-references/search?q=#1}{#1}}
\newcommand{\abr}[1]{\textsc{#1}\xspace}
\title{Cross-referencing using Fine-grained Topic Modeling}

\author{
    Jeffrey Lund, Piper Armstrong, Wilson Fearn, Stephen Cowley, Emily Hales, Kevin Seppi\\
    Computer Science Department\\
    Brigham Young University\\
   {\tt \{jefflund, piper.armstrong, wfearn,}\\
   {\tt scowley4, emilyhales, kseppi\}@byu.edu}\\
}

\date{}

\begin{document}
\maketitle

\label{ch:xref}

\begin{abstract}
Cross-referencing,
which links passages of text to other related passages,
can be a valuable study aid for facilitating comprehension of a text.
However, cross-referencing requires first, a comprehensive thematic knowledge of the entire corpus,
and second, a focused search through the corpus specifically to find such useful connections.
Due to this, cross-reference resources are prohibitively expensive and exist only for the most well-studied texts (e.g. religious texts).
We develop a topic-based system for automatically producing candidate 
cross-references which can be easily verified by human annotators. 
Our system utilizes fine-grained topic modeling with thousands 
of highly nuanced and specific topics to identify verse pairs 
which are topically related. 
We demonstrate that our system can be cost effective compared 
to having annotators acquire the expertise necessary to produce 
cross-reference resources unaided.
\end{abstract}

\section{Introduction}
\label{xr:sec:intro}

Cross-references are references within or between bodies of text
that can help elaborate upon or clarify that text.
They can be a useful tool for deep
understanding and can also be used to analyze
the relational structure of a text.
In contrast to a word concordance which simply shows passages which 
share a common keyword, cross-references often include links
which do not necessarily share the same keywords,
but are still related topically.
The existence of a thorough and complete cross-reference resource can
facilitate better scholarship of a text
and help readers to quickly find clarifying information
or see repeated themes throughout a text.
%Could use philosophy example where we might see the different times that Socrates talks about 
%the good in different works by Plato

Compared to language tasks such as part-of-speech tagging, producing
cross-reference annotations is much more labor intensive.
To produce cross-references,
annotators must become intimately familiar with a text in order
to note that a particular passage is related to another passage they happen to
recall.
This level of familiarity and expertise with a particular text typically requires
the annotator to spend a great deal of time studying and reading.
Possibly for this reason, expansive cross-references
have only been produced for the
most well-studied texts, such as the Bible.
Other texts, such as academic textbooks, may include indices or other similar
references, but these tend to be sparse, focusing on a small number of
keywords rather than linking each individual passage with other relevant
passages.

The process of creating such a resource can be
expensive and time consuming.
For example, the Bible%
\footnote{\url{https://www.lds.org/scriptures/bible}}
published by The Church of Jesus Christ of Latter-day Saints
includes numerous cross-references and topic-based categories.
These cross-references 
took hundreds of volunteers thousands of hours over seven years to produce~\cite{lds-xref}.
This process involved collecting more than 19,900 manually curated
entries from volunteers, and then editing and refining those references with a
small committee of experts down to a final cross-reference database containing
12,475 entries.

% In addition being more labor intensive that part-of-speech tagging,
% cross-referencing also grows quadratically with the size of the data.

Cross-referencing grows quadratically with the size of the data because for each of $n$ passages,
there are $n-1$ possibly related passages, yielding $\mathcal{O}(n^2)$ potential pairs.
This differs from tasks such as part-of-speech
tagging where 
annotators can tag individual sentences in isolation ($\mathcal{O}(n)$).
We can, however, evaluate pairs in isolation.
Therefore, our approach is to produce a system which utilizes fine-grained
topic modeling in order to dramatically lower the cost of producing a
cross-reference resource for new texts.
We do not expect that such a system will produce only--or even primarily--valid cross-references,
but we hope that the system could be accurate enough to allow annotators to simply
review the proposed cross-references and reduce the search cost.

% Therefore, we propose to find a manageable number of likely candidate pairs
% and then ask 
% annotators to evaluate whether these pairs are in fact related.

% It is similar in that pairs can be shown individually 
% 
% Cross-referencing differs from tasks such as part-of-speech tagging in other important ways.
% Annotators can be shown individual sentences,
% and the annotation can be made
% with no familiarity with outside passages.
% We can ask annotators to evaluate individual cross-references by
% showing them two short passages and asking them if the two passages are in fact
% related,
% but because the number of potential cross-references grows quadratically with
% the size of the data,
% we need a way to filter potential cross-references to a manageable size.

\begin{figure*}[!b]
\centering
\includegraphics[width=.9\linewidth]{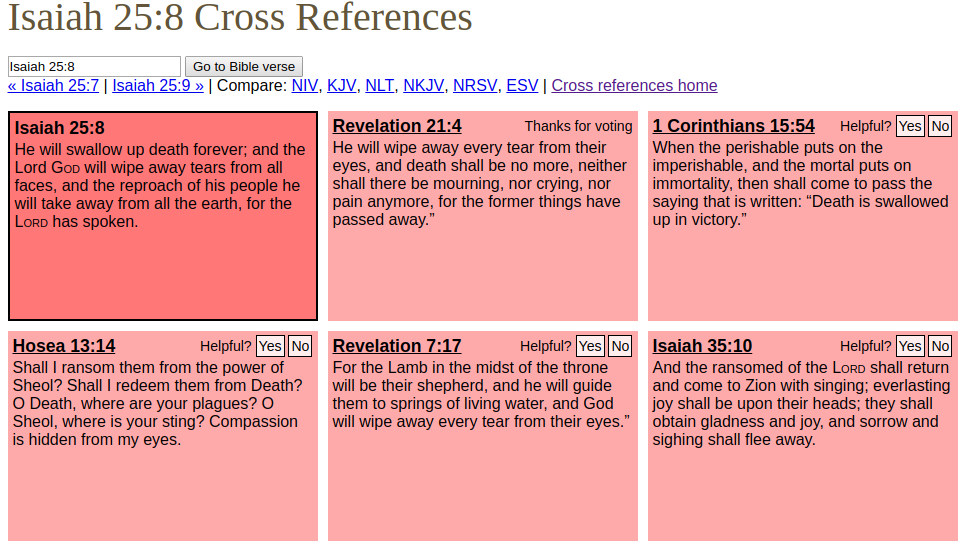}
\caption{Voting interface for cross-reference data from \url{openbible.info}.}
\label{xr:fig:openbible}
\end{figure*}

\section{Methodology}
\label{xr:sec:method}

In this section, we describe our experimental setup
and how we approach the problem of automatic cross-reference
generation.

\subsection{Cross-reference Datasets}
\label{xr:ssec:data}

While exact statistics are impossible to obtain, the number of printed copies
of the Bible is estimated to be more than 5 billion~\cite{bible-stats}.
% Because of the way Christians view this text,
It is one of the most well-studied texts in existence,
and one of the few texts in the world with extensive cross-reference data.
We utilize the English Standard Version of the Holy Bible~\cite{esv} in order to validate our method.
We use this specific translation of the Bible
because it is used on \url{openbible.info}.
While our work focuses on this specific religious text out of necessity,
it can also be applied to other
texts, including literary classics and collections of historical documents.

To aid any readers who are unfamiliar with this religious text,
we note that the term `verse' refers to a short division of a chapter.
The entirety of the text is divided up into books.
Typically, individual verses are referenced by the name of the book,
the chapter number, and the verse number, e.g., \obverse{Isaiah 25:8}.
For convenience, each referenced verse in electronic versions of this
paper is a hyperlink to \url{openbible.info}, showing the verse along
with associated cross-references.
For the sake of narrowing our focus,
our efforts in cross-referencing the Bible will focus on finding
topically related verses, though other work could potentially link
larger passages, such as entire chapters or passages spanning multiple verses.

As a ground truth for cross-references, we utilize two sources.
The first is the ``Treasury of Scripture Knowledge, Enhanced''~\cite{tske}
(an extended version of the original ``Treasury of Scripture
Knowledge''~\cite{tsk}), which includes 670,796 cross-references between
the 31,085 verses of the Bible.
To our knowledge, this is the most exhaustive resource of human curated
cross-references for the Bible to date.
We will denote this cross-reference dataset as \textit{TSKE}.

The second source of ground truth cross-references is a dataset
from \url{openbible.info}.
This dataset was seeded with various public domain cross-reference data,
including the Treasury of Scripture Knowledge.
As shown in Figure~\ref{xr:fig:openbible},
users search for a verse they are interested in
and can then vote on whether they found a particular
cross-reference to be helpful or not.
With each helpful or not helpful vote counting as $+1$ and $-1$ respectively,
the dataset includes the net result of the votes for each included
cross-reference.

Thus we can filter the dataset of cross-references based
on how helpful each verse was rated to be.
Counting only those cross-references which have a non-negative vote total,
this dataset contains 344,441 cross-references.
In figures, we denote this subset of the \url{openbible.net} cross-reference
dataset as \textit{OpenBible+0}.
We also use the subset of cross-references which received a net total
of at least 5 helpful votes.
This subset, denoted as \textit{OpenBible+5}, has 50,098 cross-references.

We do note however that the voting data has some skewness
in the number of votes for each cross-reference.
The overwhelming majority of cross-references received fewer than five total
votes for or against the reference.
A small number of verses,
including both popular verses as well as verses which happen to come
from the very beginning of the Bible,
have received hundreds of votes.

\subsection{Baselines for Automated Cross-reference Generation}

These cross-reference datasets were produced at a tremendous cost in time and
human effort.
To the best of our knowledge, efforts at automating this process are 
limited and have not received much attention in computer science literature.
That said, a reasonable baseline for automated efforts is a simple word-based
concordance, which lists words along with references to where the words occur
in the text%
\footnote{For an example of such a concordance for the King James Version of the
Bible, see Strong's Concordance~\cite{strongs-concordance}.}.

Using the \abr{TSKE} as the ground truth for cross-references,
this simple baseline will recover roughly $65\%$ of the cross-references.
For example, as shown in Figure~\ref{xr:fig:openbible}
and assuming that stemming is performed,
the verse \obverse{Isaiah 25:8} would be properly linked to
\obverse{1 Corinthians 15:54} due to the two verses
sharing the terms `death' and `swallow'.
On the other hand, verses such as \obverse{Hosea 13:14} or
\obverse{1 Corinthians 15:55}
which reference the `sting of death'
%in \obverse{Isaiah 25:8}
should not be linked
to verses such as \obverse{Revelation 9:10},
which references the sting of a scorpion.
For this reason,
roughly $99\%$ of cross-references found using word-based concordance
are spurious according to the \abr{TSKE},
making this baseline less useful as a cross-referencing resource.%
\footnote{For this reason,
published biblical word concordances typically only give a
manually curated subset of significant vocabulary terms.}
We refer to this baseline as \textit{word match}.
%Should we include here or below what the data points on the graphs mean with relation to these
%baselines? Because I'm unclear as to their meaning.

As a slightly stronger baseline,
we also consider a topical concordance in which verses
assigned to the same topic by some topic model are considered to be linked.
We refer to this baseline as \textit{topic match}.
For example, suppose that a topic model includes a topic 
which gives high probability to terms such as
`death', `swallow', `victory' and `sting'.
Assuming that such a model would assign the previously mentioned verse to this topic,
then verses such as
\obverse{Isaiah 25:8}, \obverse{Hosea 13:14}, and \obverse{1 Corinthians 14:54} would be linked,
but \obverse{Revelation 9:10} which uses the term `sting' in a different
context (i.e., the sting of a scorpion) would not be linked.

We can further increase the precision of this baseline by only linking references
which share a topic and a word, although this does come at the cost of recall.
We refer to this final baseline method as \textit{topic-word match}.

\subsection{Topic-based Cross-referencing}
\label{xr:ssec:xref-metrics}

We now describe our approach to topic-based cross-referencing.
The baselines built upon word or topic concordances
simply propose any cross-reference
for which a word or topic matches another verse,
meaning that we cannot set a threshold on the quality of the proposed
cross-references.
Instead, we propose comparing document-topic distributions as $K$-dimensional
vectors, where $K$ is the number of topics, using standard vector distance metrics to compare verses.
This idea has been used before~\cite{topic-sim1}, although not for the task
of producing cross-references.
By using a vector distance metric to compare the topical similarity
of verse pairs,
we can set a threshold on the number of proposed cross-references
and propose only the most topically related verse pairs as cross-references.
We experiment with four distance metrics:
cosine distance, Euclidean distance, cityblock (or Manhattan) distance, and Chebyshev distance.
However, given that previous work comparing document-topic vectors from LDA
seem to default to cosine similarity~\cite{topic-sim1,topic-sim2},
we anticipate that cosine distance will be the best metric for
selecting cross-references.

\subsection{Model Selection}
\label{xr:ssec:xref-models}

We claim that a fine-grained topic model,
i.e., a topic model with a large number of highly nuanced topics,
will be able to provide more value for tasks like
cross-referencing than traditional coarse-grained topic models.
In order to validate this claim, we will compare our fine-grained models
with topics from a traditional Latent Dirichlet Allocation model with 100 topics.
We refer to this baseline model as \textit{coarse}.

Traditional probabilistic topic models such as Latent Dirichlet Allocation are not
able to utilize large numbers of topics~\cite{priors-matter}.
However, we successfully train anchor-based topic models with thousands of topics.
Consequently, for our fine-grained models, we will employ the
Anchor Word algorithm~\cite{anchors-practical}.
Anchor-based topic models view topic modeling as non-negative matrix factorization.
This class of topic models attempts to decompose a document-word matrix
into two matrices, including a topic-word matrix which gives the conditional
probabilities of a particular word given a topic.
Ordinarily, this factorization is NP-Hard~\cite{beyond-svd}.
However, given a set of anchor words,
or words which uniquely identify a topic,
the computation requires only $\mathcal{O}(KV^2 + K^2VI)$
where $K$ is the number of topics, $V$ is the size of the vocabulary,
and $I$ (typically around 100)
is the average number of iterations ~\cite{anchors-practical}.

We train our anchor-based model using 3,000 topics.
We choose this number based on the number of documents we expect
each topic to explain:
there are roughly 30,000 verses and, according to OpenBible+0,
a median of 10 cross-references per verse,
so we want each topic to be responsible for roughly 10 documents.

By default, the anchors for the 3,000 topics are produced using a
modified form of the Gram-Schmidt process~\cite{anchors-practical}.
This process views each word as a vector in high-dimensional
space and attempts to pick anchor words which maximally span that space.
For more details, see~\citet{anchors-practical}.
In our results and figures,
we refer to this model with the default anchor selection method as
\textit{Gram-Schmidt}.

This does present us with some difficulty with anchor words as this
process tends to
select the most extreme and esoteric anchors possible~\cite{anchors-tsne},
which can lead to less useful topics as we increase their number.
~\citet{tandem-anchors} introduced a method of using multiple words
to form a single anchor.
This method, called tandem anchoring,
was originally formulated as a way to extend the
anchor algorithm to allow for interactive topic modeling.
%However, in the case of fine-grained topic modeling,
%the algorithm is not scalable
%enough to allow interaction with such a large number of topics.%
%\footnote{With a typical number of topics, say 50, it takes less than 3 seconds to update
%on an \abr{amd} Phemon II X6 1090T processor. In contrast, with 3,000 topics, topic recovery
%takes approximately 2 hours and 20 minutes.
%This can be improved to a few minutes with proper parallelization
%but even this is not fast enough for an interactive system.}

Instead of utilizing human interaction to seed the topic anchors,
we will seed the tandem anchors using the terms from
randomly selected verses.
For example, suppose we randomly select \obverse{Isaiah 25:8} as 
a verse from which to form an anchor.
As shown in Figure~\ref{xr:fig:openbible}, this verse
includes terms such as `swallow', `death', and `tears'.
Each of these terms is represented as a point in high-dimensional space.
To produce a single anchor from these terms,
we average the words using the element-wise harmonic mean%
\footnote{See \citet{tandem-anchors} for details on why
the harmonic mean is useful for forming tandem anchors}.
While this new point may not correspond to any particular word,
it does capture the joint occurrence pattern of the words which
form the anchor.
We repeat this process 3,000 times to produce an anchor-based topic
model with tandem anchors.
While this exact methodology of seeding topic anchors using
randomly selected verses is novel,
we note the similarity to the method used to seed topics in Rephil,
a web scale topic model used by Google~\cite{murphy-rephil}.
In figures, we refer to this model with tandem anchors as \textit{tandem}.

For each of these models, we must take the topic-word distributions
from the topic model and produce document specific topic assignments.
We utilize mean field variational inference in order to assign the
individual verses to topics,
similar to~\citet{supervised-anchors}.

\begin{figure*}[b!]
\centering
\resizebox{1\textwidth}{!}{\input{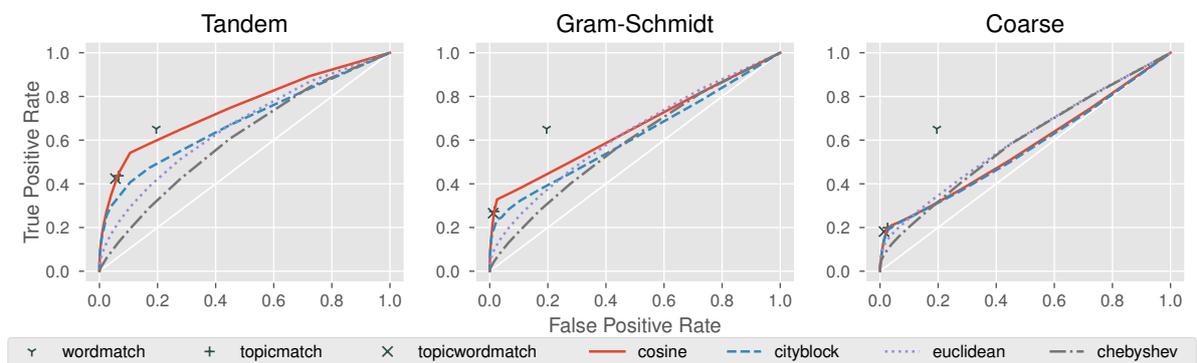}}
\caption{Cross-reference ROC curves for different metrics for cross-reference
selection with topics from three different topic models
and TSKE as the cross-reference ground truth.}
\label{xr:fig:roc_metric}
\end{figure*}

\begin{figure*}[b!]
\centering
\resizebox{1\textwidth}{!}{\input{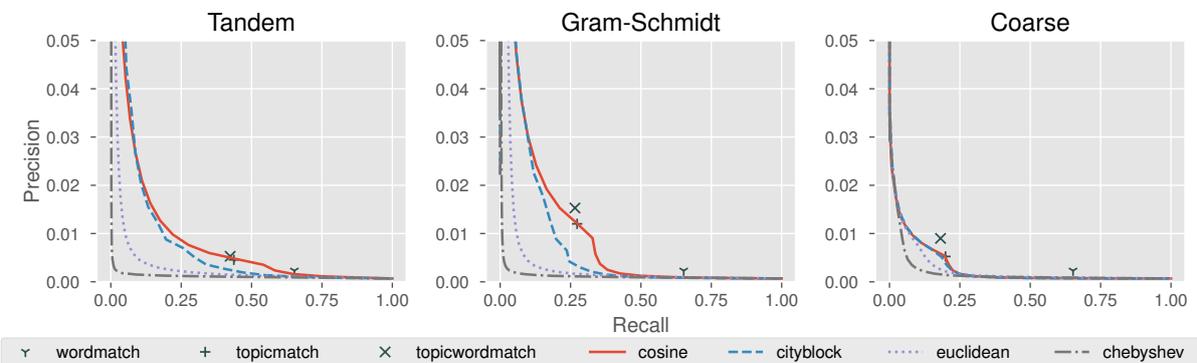}}
\caption{Cross-reference PRC plots with different metrics for cross-reference selection
with topics from three different topic models
and TSKE as the cross-reference ground truth.}
\label{xr:fig:prc_metric}
\end{figure*}

\section{Results}

In this section we present the results of our experiments 
with topic-based cross-referencing.
As discussed in Section~\ref{xr:sec:method},
we experiment with three different cross-reference datasets:
TSKE, OpenBible+0, and OpenBible+5.
We utilize three different topic models:
coarse, Gram-Schmidt, and tandem.
We seek to demonstrate that our topic-based cross-referencing system
can effectively utilize fine-grained topic modeling to
produce candidate cross-references which can be annotated
by humans in a cost effective manner.

\subsection{Metric Comparisons}
\label{xr:ssec:results-metrics}

We first explore the various metrics for selecting cross-references
discussed in Section~\ref{xr:ssec:xref-metrics}.
With each proposed distance metric,
we are able to set a threshold and determine which Bible verse pairs
to keep as candidate cross-references and which to discard.
Figure~\ref{xr:fig:roc_metric} and Figure~\ref{xr:fig:prc_metric} summarize
our results with respect to metrics.

Figure~\ref{xr:fig:roc_metric} gives a receiver operator characteristic (or ROC) curve,
which compares the true positive rate (or recall) against the
false positive rate (or fall-out).
%Should we have something here about what this tells us?
%I.E. why do we want to use a ROC curve?
We show the curve for the various metrics using
the TSKE as our ground truth.
We show these curves on each of the three 
different models discussed in
Section~\ref{xr:ssec:xref-models}.

Overall, cosine distance is the best method for selecting cross-references,
as it gives the largest area under the ROC curve.
The major exception to this is with the traditional coarse-grained topic model
for which Euclidean distance performs the best.
Considering that cosine distance has frequently been used
in conjunction with topics from LDA~\cite{topic-sim1,topic-sim2},
this result is somewhat surprising.

Also of interest is the fact that the word-match baseline does
reasonably well with respect
to the true positive rate, at least if a false positive rate of roughly $0.196$ is acceptable.
Note that this corresponds to 188,974,806 false positives in the \abr{TSKE} dataset, so while the
raw number of true positives may be impressive,
this baseline is not likely to be useful in practice.

\begin{figure*}[b!]
\centering
\resizebox{1\textwidth}{!}{\input{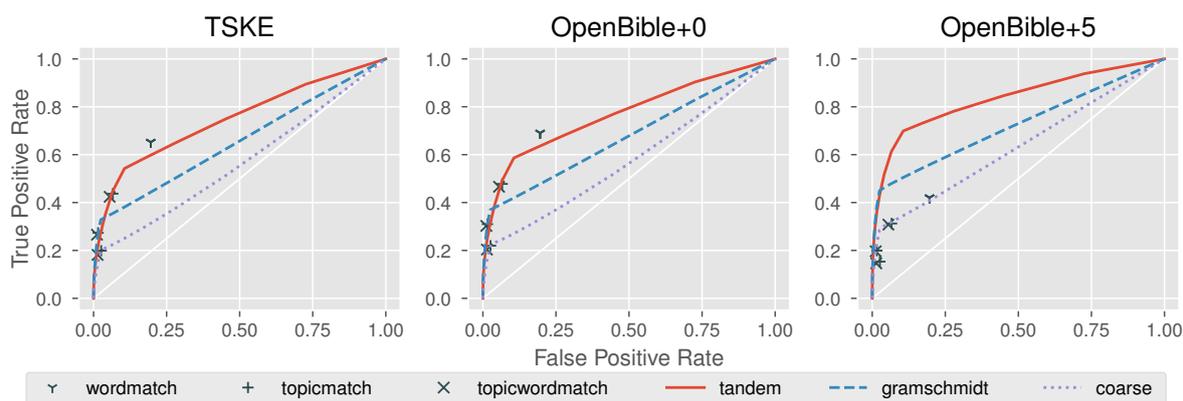}}
\caption{Cross-reference ROC curves with different models for cross-reference
selection with three different datasets.}
\label{xr:fig:roc_models}
\end{figure*}

\begin{figure*}
\centering
\resizebox{1\textwidth}{!}{\input{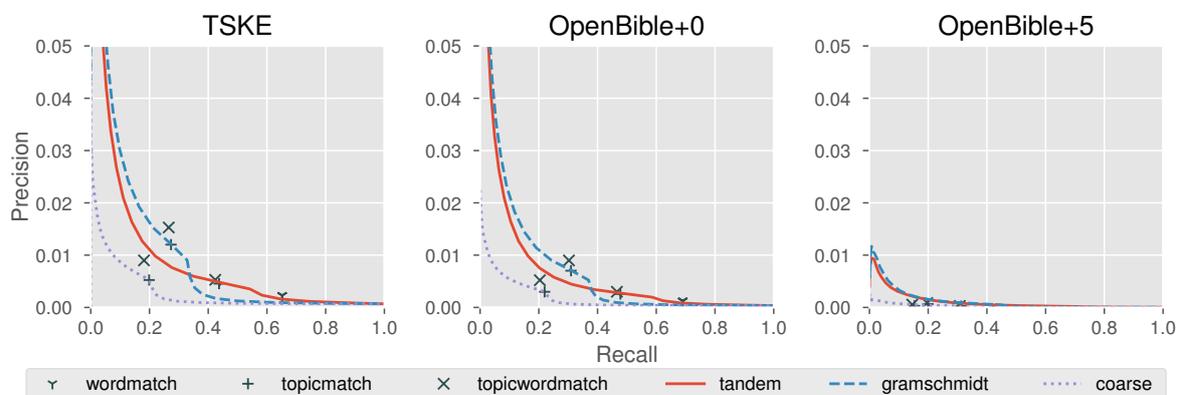}}
\caption{Cross-reference PRC plots with different models for cross-reference
selection with three different datasets.}
\label{xr:fig:prc_models}
\end{figure*}

While the ROC plot is undoubtedly more popular than
the precision-recall plot, in cases where the data is imbalanced,
the precision-recall plot can be much more informative.
This is mainly because of the use of false-positives in the ROC curve,
which can
present an overly optimistic picture of the classifier
performance~\cite{prc-plots}.
Consequently, in Figure~\ref{xr:fig:prc_metric}, we also compare each of the proposed
metrics using a precision-recall curve (or PRC).

The PRC plot reinforces the claim that cosine distance is the 
best distance metric
to threshold cross-references
since for any reasonable level of precision,
cosine distance yields the best results.
Once again, with the exception of Euclidean distance
performing better with the coarse-grained topic model.
However, we note that the precision using coarse-grained topics
is much lower than using fine-grained topics with cosine distance.
Even with Gram-Schmidt based fine-grained topics,
where Euclidean distance
eventually wins out against cosine distance,
the high amount of false positives
means that cosine distance is the most useful metric for this task.
The PRC plot also illustrates why the matching baselines are not
practically useful---they do have decent true positive rates, but the precision
with these baselines is extremely low.

\subsection{Topic Model Comparison}

We now explore the various topic models discussed in 
Section~\ref{xr:ssec:xref-models}.
Figure~\ref{xr:fig:roc_models} and Figure~\ref{xr:fig:prc_models}
summarize these results.
Based on Section~\ref{xr:ssec:results-metrics},
each reported result in this section uses
cosine distance to determine verse pairs which should be
considered as candidate cross-references.
We compare the results of the three topic models on the
three datasets discussed in Section~\ref{xr:ssec:data}.

\begin{figure*}
\centering
\resizebox{1\textwidth}{!}{\input{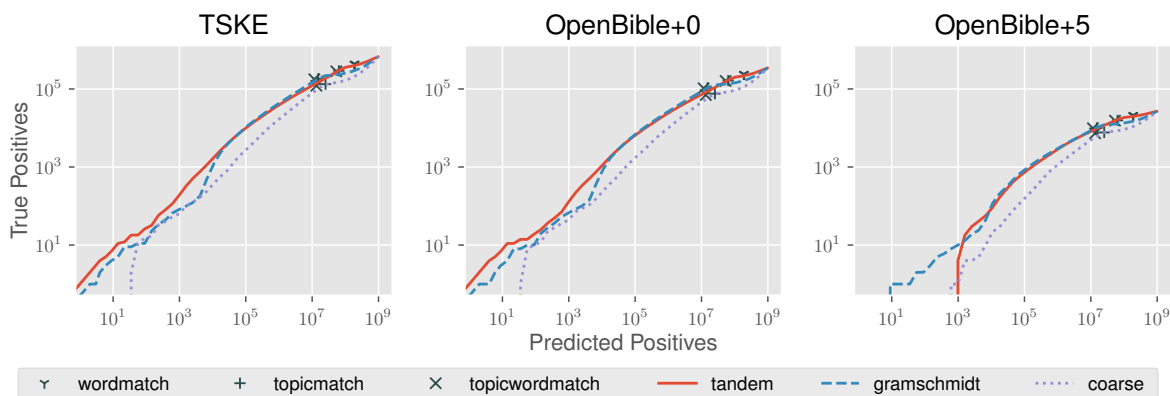}}
\caption{Cross-reference cost curves with different models for cross-reference selection
with three different datasets.
Note the log-log scale.
The x-axis denotes the number of cross-references produced by our method,
while the y-axis indicates how many of those cross-references are
valid according to the human-provided ground truth.
}
\label{xr:fig:cost}
\end{figure*}

The overall trend on each of the three datasets
is simply that we lower the true positive rate
and precision as we use more selective datasets.
Considering that OpenBible+5 is a subset of OpenBible+0,
and that OpenBible+0 is nearly a subset of TSKE%
\footnote{OpenBible+0 has 533 references seeded from other public domain sources,
which are not included in TSKE.},
this result is not surprising.
In the final cost analysis (see Section~\ref{xr:ssec:cost}),
the selectivity of the actual annotators
must be taken into account when attempting to predict
the true positive rate or the precision of a system.

As shown in Figure~\ref{xr:fig:roc_models},
regardless of the anchor selection strategy, both fine-grained topic models outperform
the traditional coarse-grained model.
Our model using tandem anchors built from randomly selected verses
performs the best
for nearly all levels of false positive rate.
However, the Gram-Schmidt based anchors produce better
true positive rates for very low
false positive rates.

This trend is better illustrated with the PRC plots in Figure~\ref{xr:fig:prc_models}.
While for higher values of recall the tandem anchor selection strategy
does win out,
it is only after precision significantly drops that
tandem anchors produce superior
predictions to Gram-Schmidt anchors.

\subsection{Cost Analysis}
\label{xr:ssec:cost}
\begin{table*}[htb!]
    \begin{tabular}{p{.35\textwidth} p{.55\textwidth}}
     \hline\hline
     \textbf{Passage 1} & \textbf{Passage 2} \\
     ``I have an excessive regard for Miss Jane Bennet, she is really a very sweet girl, and I wish with all my heart she were well settled. But with such a father and mother, and such low connections, I am afraid there is no chance of it.'' & 
    ``I wish you may not get into a scrape, Harriet, whenever he does marry;--I mean, as to being acquainted with his wife--for though his sisters, from a superior education, are not to be altogether objected to, it does not follow that he might marry any body at all fit for you to notice. The misfortune of your birth ought to make you particularly careful as to your associates. There can be no doubt of your being a gentleman's daughter, and you must support your claim to that station by every thing within your own power, or there will be plenty of people who would take pleasure in degrading you.''\\ 
     \hline
    \end{tabular}
    \caption{Passage 1 is taken from the eighth chapter of \textit{Pride and Prejudice}, and Passage 2 is taken from the fourth chapter of \textit{Emma}. These two passages are a valid cross-reference because they both discuss social standing and family connections in the context of marriage. Their connection was found even with their lack of shared words.}
    \label{xr:tab:austen}
\end{table*}

While the precision-recall curves in Figure~\ref{xr:fig:prc_models}
may suggest that
Gram-Schmidt based topic models produce superior topics for cross-referencing,
we suggest that this analysis may be missing a key point in real world
analysis.
As an alternative to both PRC and ROC curves,
we suggest that this task might be
best served with an analysis of cost per true positive.

We envision that our system would be used to produce a set of candidate
cross-references which would then be curated using human annotators.
These annotators
would be tasked with evaluating each potential reference
and determining whether or not each cross-reference is valid.
Critically, the annotator would only be required to evaluate
individual cross-references, not the entire text.

As a working example of the cost of such an annotation process,
suppose we use a popular
crowd-sourcing service (e.g., Amazon Mechanical Turk) to produce
human annotations.
We might reasonably expect to pay something around
\$0.01 USD per annotation.
We would likely require some form of quality control in the
form of redundant annotations,
so we might end up paying \$0.05 USD
per annotated cross-reference candidate.
Of course,
the exact cost per annotated cross-reference will vary depending
on the service and difficulty of the specific text being
cross-referenced.
However, we will use these estimates for the purpose
of illustration.

Suppose as part of this working example,
we are interested in producing a resource
with 12,000 valid cross-reference annotations
(roughly matching the size of the previously mentioned
LDS edition of the Bible~\cite{lds-xref}).
Consulting Figure~\ref{xr:fig:cost} we can then determine how many
candidate cross-references
we would need to produce for human annotation in order to create
the final curated cross-reference resource.
For example, using the TSKE as our ground truth,
we would need approximately 150,000 predicted positives
in order to find 12,000 true positives.
At \$0.05 USD per annotation, this would cost about \$7,500 USD.
Supposing that our annotators were more selective,
we could use the OpenBible+0 as the ground truth,
which would roughly double the cost.
With OpenBible+5, which is considerably more selective,
this cost rises to approximately \$1,000,000 USD.
In contrast, with traditional coarse-grained topic modeling,
this cost is anywhere from \$40,000 USD using the TSKE as the ground truth,
to \$17,500,000 USD using OpenBible+5 as the ground truth.

\vspace{-1cm}
While these costs may seem prohibitive, consider that the alternative is to
have experts understand the entire text to the degree that they can read
one passage and recall other relevant passages they have previously read.
In the case of religious texts, this is often possible since adherents
study those texts as part of their daily routine.
For example, in the case of the LDS edition, it took a committee of experts
seven years of work to produce their cross-reference resource, even with the
aid of hundreds of volunteers.
However, without those experts and volunteers, the cost would have been
even greater.
In the naive case where every possible reference is manually checked,
the cost skyrockets to around \$48,000,000 USD.

\section{Discussion}
Without extensive cross-referencing resources for more secular datasets,
it is difficult to empirically prove the usefulness of our system generally
without an extremely costly user study.
That said, we make a small attempt by manually examining cross-references generated
from the complete works of Jane Austen and Plato.
Based on our cost analysis in Section~\ref{xr:ssec:cost},
and since we will be examining only a small number of cross-references,
we utilize tandem anchors to generate topics.
With each dataset, we examine the first 300 cross-references
produced by our system.

We also examine what our model got wrong in proposing Bible cross-references.

\subsection{Jane Austen}

Of the first 300 cross-references, we find that 39 of them are valid,
linking passages from all six works by Jane Austen.
As with our experiments with the Bible,
this level of precision is sufficient that we believe
that we could dramatically lower the cost of producing a full
cross-referencing resource for this text.

We note that 109 of these are cross-references linking a paragraph in
the eighth chapter of \textit{Pride and Prejudice} to other
passages in our corpus.
Of the references involving this one paragraph, 22 were valid.
An example of such a cross-reference is shown in Table~\ref{xr:tab:austen}.
While marriage in general is a common theme in the works of Jane Austen,
this particular paragraph more specifically discusses the role of social status and family connection
as it relates to choosing a marriage partner.
We note that the connection between the passages in Table~\ref{xr:tab:austen}
is thematic;
they share no significant words in common,
demonstrating the capability of the system to detect nuanced topics and themes.

\subsection{Plato}

Of the first 300 cross-references generated from the works of Plato, we found 119
that were valid.
Many of the cross references were between distinct works, and included discussions
about the nature of justice, arguments about the composition of things,
the nature and role that certain things play, and discussions
of appropriate legislation.

It is important to note that the model found significantly more cross references
in the works of Plato than those by Jane Austen.
This is likely due to the nature of the writing.
We find in the works of Plato that ideas themselves are discussed directly, similar
to the bible, and thus we would expect it to be easier for a model to find
words and phrases that link to a specific topic.
An example of this is that words like ``virtue," ``courage," ``mean," and ``cowardice"
would likely identify a topic about virtue that comes up in the works of Plato.
However, in Jane Austen we find ideas discussed implicitly through interactions of the characters
and commentary by the author.
The meaning, while present, is found by ``reading between the lines.''
An example of this is that we might find marriage discussed in the works of Jane Austen, but
more often through characters discussing their feelings after getting married.
Here, words that we might see could be words such as ``happy," ``elated," ``love," ``efficient," etc.
However, these words could also correlate equally well with other topics, and thus it
would be harder for our model to discern.
As further evidence of this we point out that the cross reference we used above from Jane
Austen is an explicit discussion of marriage.
It is likely that an implicit discussion of marriage would be harder for our model to find.
We also point out that in such cases it is a non-trivial task for humans to come to a specific
consensus about what a given passage could mean or relate to.

That said, given the relevant and influential nature that the works of Plato still hold even
today, we can see that these cross references are highly useful in that they
facilitate study and understanding of his works that a study of each individual work
separately might miss.

\subsection{Error Analysis}
We examine the errors of running tandem anchors using cosine distance on the Bible.
There are two types of errors to examine:
candidate cross-references proposed early in the process that are not valid cross-references
and valid cross-references that are not proposed until the end of the process
(as determined by the Treasury of Scripture Knowledge).

Early invalid candidate cross-references all exhibit the same characteristic;
the documents are exactly or substantively the same
(e.g. \textit{\obverse{Deuteronomy 2:17} the Lord said to me,} and \textit{\obverse{Deuteronomy 2:2} Then the Lord said to me}).
Indeed, a human given only those two documents would also mark them as related,
and many valid cross-references exhibit this same characteristic
(e.g. Psalms 107:6 and Psalms 107:28 are exactly the same and are a valid cross-reference).

Cross-references are partially so difficult because what constitutes a valid cross-reference
is at least partially determined by what the community surrounding the text views as significant,
and so two documents that are identical may or may not be a valid cross-reference
depending on the view of that community.
This particular issue would make any end-to-end automated solution to cross-referencing particularly difficult.

We also examined the last one hundred valid cross-references proposed.
We consider these errors because in a system
where a predetermined number of candidate cross-references are considered,
these candidate cross-references would most likely never be considered.
For 28 of the last one hundred valid cross-references,
it is unclear why they are considered valid cross references
(e.g., \textit{\obverse{Ezekiel 16:10} I clothed you also with embroidered cloth and shod you with fine leather.
I wrapped you in fine linen and covered you with silk.} and
\textit{\obverse{Deuteronomy 8:11} ``Take care lest you forget the Lord your God by not keeping his commandments and his rules and his statutes,
which I command you today,}).

Many of the other 72 valid cross-references are also difficult in some way. Many of the connections involve some use of metaphor (25)
or are linked by a single key word, such as a name, but are otherwise topically dissimilar (34).
The other 13 cross-references are all documents describing the construction of the tabernacle,
and we have enough extra context to recognize this,
however it isn't surprising that the model doesn't find this connection.

It is useful to note that of the final 600,000 candidate cross-references,
only one hundred of them were valid cross-references.

\section{Conclusion}
We have produced a system using fine-grained topic modeling which is able to
propose candidate cross-references which can be verified by non-expert human
annotators for the purpose of creating a cross-reference resource at a fraction
of the cost of current manual techniques.
Our method, which utilizes tandem anchors to produce large numbers of 
highly nuanced topics coupled with an effective assignment strategy, is able
to produce document-topic vectors which are comparable using cosine distance.

Our results also demonstrate that this system would not be as cost effective
with traditional coarse-grained topic modeling.
While we can find sets of topically related documents using coarse-grained
topics,
for the task of finding the most closely related documents we require a
system which is more specific.
We suggest that this success serves as motivation for exploration of
fine-grained topic modeling for other topic-based use cases which
require nuance and precision.

\section*{Acknowledgements} This work was supported by the NSF Grant IIS-1409739
\bibliographystyle{acl_natbib}
\bibliography{main}

\end{document}